\documentclass[conference]{IEEEtran}

\IEEEoverridecommandlockouts
\usepackage{cite}
\usepackage{amsmath,amssymb,amsfonts}
\usepackage{algorithm}
\usepackage{float}
\usepackage{algorithmic}

\usepackage{url}
\usepackage{hyperref}
\usepackage{graphicx}
\usepackage{textcomp}
\usepackage{stfloats}
\usepackage{xcolor}
\usepackage{booktabs}
\usepackage{multirow}
\usepackage{url}

\def\BibTeX{{\rm B\kern-.05em{\sc i\kern-.025em b}\kern-.08em
    T\kern-.1667em\lower.7ex\hbox{E}\kern-.125emX}}

\begin{document}

\title{Winning Solution to UrbanTwin Sim2Real LiDAR Challenge 2026 (V2X-Real Track): Site-Calibrated Synthetic Data and Multi-Detector Fusion}

\author{\IEEEauthorblockN{Arnav Kumar}
\IEEEauthorblockA{\textit{Department of Computer Science and Engineering} \\
\textit{The Ohio State University}\\
Columbus, OH, USA \\
kumar.1178@osu.edu}
\and
\IEEEauthorblockN{Brijesh Soni}
\IEEEauthorblockA{\textit{Department of Computer Science and Engineering} \\
\textit{The Ohio State University}\\
Columbus, OH, USA \\
soni.152@osu.edu}
}

\maketitle
\thispagestyle{empty}
\pagestyle{empty}
\begin{abstract}
This work summarizes the winning solution for the UrbanTwin Sim2Real LiDAR Challenge (V2X-Real track) at the IEEE BigData Cup 2026. The primary objective is to train 3D detectors for vehicles, pedestrians, and trucks using \textit{only synthetic} roadside LiDAR data, and to generate synthetic scans that are sufficiently realistic to be indistinguishable from real ones. A CARLA-based pipeline was developed and calibrated to the fixed sensor site, accurately modeling the range image, including no-return beams, based on unlabeled real development scans. A combined inference detector, utilizing multiple 3D detectors (VoxelNeXt, PV-RCNN, and SECOND) trained exclusively on synthetic data, was also proposed. Each scan was aligned to the synthetic ground plane and processed through the detectors at multiple ground heights. The resulting bounding boxes were fused using bird’s-eye view (BEV) clustering, followed by class-specific post-processing to refine heading, box size, and filter low-confidence detections. The final submission achieved a joint rank 1 with a combined score of 0.8947. The ablation study demonstrated that per-scan ground leveling more than doubled the real-scan mAP, while fusing boxes across different heights and models further improved performance. The proposed work closes the gap between synthetic and real LiDAR data for 3D object detection in Urban V2X scenarios. Reproducible code available at \url{https://github.com/Arnavvvkumar/IEEE-BigData-2026-Cup---V2X-Track}

\end{abstract}


\section{Introduction}
\label{sec:intro}


Developing a roadside LiDAR detector involves more than simply installing a sensor at an intersection. It requires annotated point clouds that precisely represent the site geometry, sensor perspective, and trajectories of road users. Generating these datasets at scale is both costly and labor-intensive, particularly when each object must be assigned an accurate 3D bounding box and class label.

To address this challenge, synthetic datasets are employed to generate labeled scenes under controlled conditions, enabling systematic variation in traffic, object placement, and sensor configuration. However, a central question raised by the challenge is: \textbf{Can synthetic LiDAR data be made realistic enough that a detector trained on it generalizes effectively to real-world frames?}
This question motivates the UCF UrbanTwin Sim2Real LiDAR Challenge \cite{ref_challenge} and forms the foundation of this study.

Realistic transfer from synthetic data to real roadside scans requires that generated point clouds accurately reflect the conditions present during real scan collection. The position and orientation of the LiDAR sensor determine the intersection perspective, while the beam pattern and range influence the spatial distribution of measurements. Collectively, these factors contribute to the simulation-to-real domain gap, which can result in discrepancies in spatial distributions between synthetic and real point clouds.
Additionally, variations in object appearance can reduce detection accuracy \cite{shahbaz2026high}. This effect is especially pronounced for pedestrians and distant road users, who may be represented by only a few points in the point cloud and therefore require
precisely localized bounding boxes. Addressing the simulation-to-real domain gap requires careful consideration of both the roadside environment and sensor characteristics.

One approach to modeling these site and sensor characteristics is the use of digital twins, which replicate features of a real
environment, including road layout, surrounding infrastructure, and
sensor placement within a simulation. LiDAR scans generated from these environments can be paired with object annotations and subsequently used to train perception models. Previous studies have investigated this approach \cite{shahbaz2025urbantwin,shahbaz2026high,shahbaz2026urbantwin}
for roadside LiDAR and have introduced the UrbanTwin synthetic datasets
for various perception tasks.

Our proposed approach differs in that, rather than reproducing the intersection geometry, it models the fixed sensor's specific view of the site. For each beam direction, the method estimates, using unlabeled real development scans, where the beam typically returns, at what range, and how frequently it returns no point.

We present our approach to the V2X-Real track of the UCF UrbanTwin Sim2Real LiDAR Challenge, in which we generate labeled
synthetic roadside scans using a site-calibrated pipeline and use them
to train multiple 3D object detectors. During inference, we align each real scan to the synthetic ground plane, combine the detectors’ predictions, and apply class-specific post-processing. We then evaluate the resulting detections on real roadside scans and assess the realism of the submitted synthetic scans.

Our main contributions are as follows:

\begin{itemize}
    \item Site-Calibrated CARLA-based data generation: We develop a synthetic data generation pipeline built on CARLA and a custom Generator V2. Using unlabeled development scans it models the fixed sensor's range image, including no-return beams and combines these with simulated traffic objects and their 3D annotations.
    \item A Combined Inference Framework: We train several 3D detectors (Sparsely Embedded Convolutional Detection (SECOND) \cite{yan2018second}, VoxelNeXt \cite{chen2023voxelnext}, and the Point-Voxel Region-based Convolutional Neural Network (PV-RCNN) \cite{shi2020pvrcnn}, including class-specific specialists) exclusively on synthetic data, and integrate them with a test-time pipeline comprising ground-plane leveling, multi-height inference, agreement-based box fusion, and class-specific calibration. Moreover, each detector contributes only to the classes where it is validated.
    \item Scoring-aware post-processing: Many of the pedestrians are very close to the label but too loose to reach an IoU of 0.5. Therefore, we add size-graded hedging of pedestrian boxes, and tune the submitted scans to match the real sensor's range and no-return statistics.
    \item We evaluate detection and realism on the V2X-Real track, achieving a combined score of 0.8947, and analyze which components contributed positively and where validation failed to predict the test result, particularly for trucks.
\end{itemize}


The rest of this paper is organized as follows. 
Section~\ref{sec:related} reviews related work.
Section~\ref{sec:problem} formulates the problem and describes the dataset. Section~\ref{sec:method} presents the proposed methodology. Section~\ref{sec:setup} details the experimental setup, and Section~\ref{sec:results} reports the numerical results. Section~\ref{sec:conclusion} concludes the paper.

\section{Related Work}
\label{sec:related}

The effectiveness of training 3D detectors on synthetic data is contingent upon the degree to which simulated LiDAR scans replicate real-world data. Simulators such as CARLA~\cite{dosovitskiy2017carla} are capable of generating substantial volumes of labeled data; however, their point clouds often diverge from real measurements in terms of beam pattern, dropout, and object appearance.

To address this discrepancy, LiDARsim~\cite{manivasagam2020lidarsim} constructs simulation assets directly from real scans. For roadside sensing, the UrbanTwin series~\cite{shahbaz2025urbantwin,shahbaz2026high,shahbaz2026urbantwin} develops high-fidelity digital twins of real intersections and provides synthetic roadside datasets for training and evaluation. Building on this site-specific simulation paradigm, the present work models the fixed sensor's perspective of the site, including beams that do not return any points, rather than reproducing the entire scene geometry.

Despite the use of realistic data, detectors trained in one domain frequently exhibit performance degradation when applied to another domain. Voxel-based detectors such as SECOND~\cite{yan2018second}, the fully sparse VoxelNeXt~\cite{chen2023voxelnext}, and the two-stage PV-RCNN~\cite{shi2020pvrcnn} are widely adopted for LiDAR detection. However, Wang \textit{et al.}~\cite{wang2020train} demonstrated that a significant portion of cross-dataset performance drop is attributable to differences in object size between domains. This finding underscores the importance of size-aware processing of trucks and pedestrians within the proposed pipeline. A common strategy involves combining multiple detectors. Weighted box fusion~\cite{solovyev2021wbf} has shown that averaging overlapping boxes from different detectors, rather than suppressing all but one, yields more accurate bounding boxes than any individual model. Building on this, the present work extends the fusion approach to integrate predictions across both models and ground heights, employing a circular mean for box headings, as detailed in Section~\ref{sec:inference}.


\section{Problem Formulation and Dataset}
\label{sec:problem}

\subsection{Problem Statement}

The problem statement for IEEE Big Data Cup 2026 UCF UrbanTwin Sim2Real LiDAR Challenge can be formalized as follows: Given a point cloud $P$
captured by a roadside LiDAR sensor, the task is to localize and
classify vehicles, pedestrians, and trucks. A detector $f_\theta$
maps $P$ to a variable-size set of predictions:

\begin{equation}
    f_\theta(P) = \{(b_i, c_i, s_i)\}_{i=1}^{M},
    \label{eq:detector-output}
\end{equation}

where $M$ is the number of detections for the input scan.
Each oriented 3D bounding box
$b_i = (x_i, y_i, z_i, \ell_i, w_i, h_i, \psi_i)$
is parameterized by its center $(x_i, y_i, z_i)$,
length $\ell_i$, width $w_i$, height $h_i$, and yaw $\psi_i$.
The term $c_i$ denotes the predicted class, and $s_i \in [0,1]$ denotes the confidence score.

Under the Track B training constraints, we learn $\theta$ exclusively from generated LiDAR point clouds and their associated 3D bounding-box and class annotations. The trained detector is then further evaluated on held-out real roadside LiDAR scans.

\subsection{Dataset Overview}

This study utilizes real roadside LiDAR scans from the infrastructure-focused portion of the V2X-Real dataset \cite{xiang2024v2xreal}, as well as synthetic scans generated for detector training.
The test set comprises 50 real scans without labels; the held-out annotations include 1,547 bounding boxes (470 vehicles, 953 pedestrians, and 124 trucks). Additionally, many objects are represented by only a few points, which motivates the localization-focused post-processing described in Section~\ref{sec:method}.
Furthermore, frames identified in the organizers' forbidden-frame file \textit{were excluded} from all stages of this study. For development, two screened sets comprising 150 and 350 real scans with cooperative annotations, none of which appear in the forbidden list, were utilized.
All detectors were trained exclusively on synthetic data generated for this study (Table~\ref{tab:dataset}).

The \emph{canonical} set contains simulated roadside scenes (CARLA and our Generator V2).  The \emph{site-calibrated twin} sets combine the modeled roadside background with simulated traffic objects and their 3D annotations; the \emph{small-truck} twin set resizes trucks to the dimensions of the real trucks, and further twin sets vary pedestrian size.

The Point-Voxel Region-based Convolutional Neural Network (PV-RCNN) models were trained on the first 3 sets (18,800 frames). Moreover, the SECOND and VoxelNeXt ensemble members (Section~\ref{sec:method}) were trained on these sets together with the size-varied pedestrian set.

\begin{table}[t]
\caption{Data used for detector training and evaluation}
\label{tab:dataset}
\centering
\begin{tabular}{lrl}
\hline
\textbf{Dataset} & \textbf{Frames} & \textbf{Use} \\
\hline
Canonical synthetic scenes   & 10,800 & Training \\
Site-calibrated twin scenes  & 5,000  & Training \\
Small-truck twin scenes      & 3,000  & Training \\
\hline
Merged synthetic set$^{a}$   & 18,800 & Training \\
\hline
Size-varied pedestrian twin scenes & 5,000 & Training$^{b}$ \\
\hline
Real development set 1       & 150 & Evaluation and selection \\
Real development set 2       & 350 & Evaluation and selection \\
Real detection test set      & 50  & Challenge evaluation \\
Submitted synthetic scans    & 50  & Realism evaluation \\
\hline
\end{tabular}

\smallskip
\footnotesize
$^{a}$ The merged set trained the three-class and class-specific PV-RCNN models. \\
$^{b}$ Used only by some SECOND and VoxelNeXt members.
\end{table}

\subsection{Evaluation Metric}

The challenge evaluates two components:

\begin{itemize}
    \item the accuracy of detections on
real roadside scans
    \item the realism of the submitted synthetic point
clouds
\end{itemize}

The detection is measured in terms of mAP (mean average precision) across vehicles, pedestrians and trucks at a 3D IoU (intersection over union) threshold of 0.5 with 40 point recall interpolation (R40). Results at IoU~0.7 are also reported.

In the realism evaluation, 50 synthetic frames are aggregated into a single point set, cropped to the challenge range, and subsampled to 10,000 points per set. The resulting distributions are then compared using Chamfer Distance (CD) and Maximum Mean Discrepancy (MMD).
In addition to MMD, Earth Mover's Distance (EMD) and Fr\'echet Point-cloud Distance (FPD) are also evaluated.
Each distance metric is normalized to the range $[0,1]$ using fixed minimum and maximum values specified by the organizers (CD: 2–30, MMD: 0–0.5, EMD: 0–5, FPD: 0–50).

The four normalized distance metrics are averaged to obtain the
realism score $(R)$. Because the score is computed on a single fixed 10,000-point sample, minor variations in $R$ partially reflect sampling noise, as observed in the measurements: different subsamples of the same scans differed by approximately $\pm 0.004$. The Detection metric is normalized similarly and capped at 1.

The final challenge score combines normalized detection and realism:

\begin{equation}
S = 0.6 \min\left(\frac{\mathrm{mAP}_{3D@0.5}}{0.7},\,1\right)
    + 0.4R. \label{eq:combined_score}
\end{equation}

\subsection{Use of Real Data}

All detectors were trained exclusively on synthetic data. Real data were used only as follows:
\begin{itemize}
    \item Unlabeled real development scans were utilized to calibrate the synthetic pipeline, as they contributed to correcting the sensor's range image and no-return statistics (Section~\ref{sec:synth}). These scans also facilitated the construction of a map of static structures that reduces pedestrian scores and established the range histogram for the submitted synthetic scans.
    
    \item Labeled real development scans: Two screened sets, comprising 150 and 350 frames, were exclusively used for evaluation and selection, and never for model training. These scans supported validation, model and checkpoint selection, and the determination of post-processing constants, including box-offset calibration, box-size statistics, the truck size rule, and point-count thresholds for pedestrians and trucks. In accordance with the established rules, no detector was trained or fine-tuned on real labels or on object points extracted using real boxes.
    
    \item Forbidden frames and test scans: These frames were completely excluded from the aforementioned uses. In compliance with competition rules, each test scan is processed independently, with its ground plane fitted using only its own points. No tuning is performed on the test scans, and neither pseudo-labels nor self-training are employed.
\end{itemize}
\section{Proposed Methodology}
\label{sec:method}

\subsection{Overview}
\begin{figure*}[t!]
    \centering
    \includegraphics[
        scale=0.23
    ]{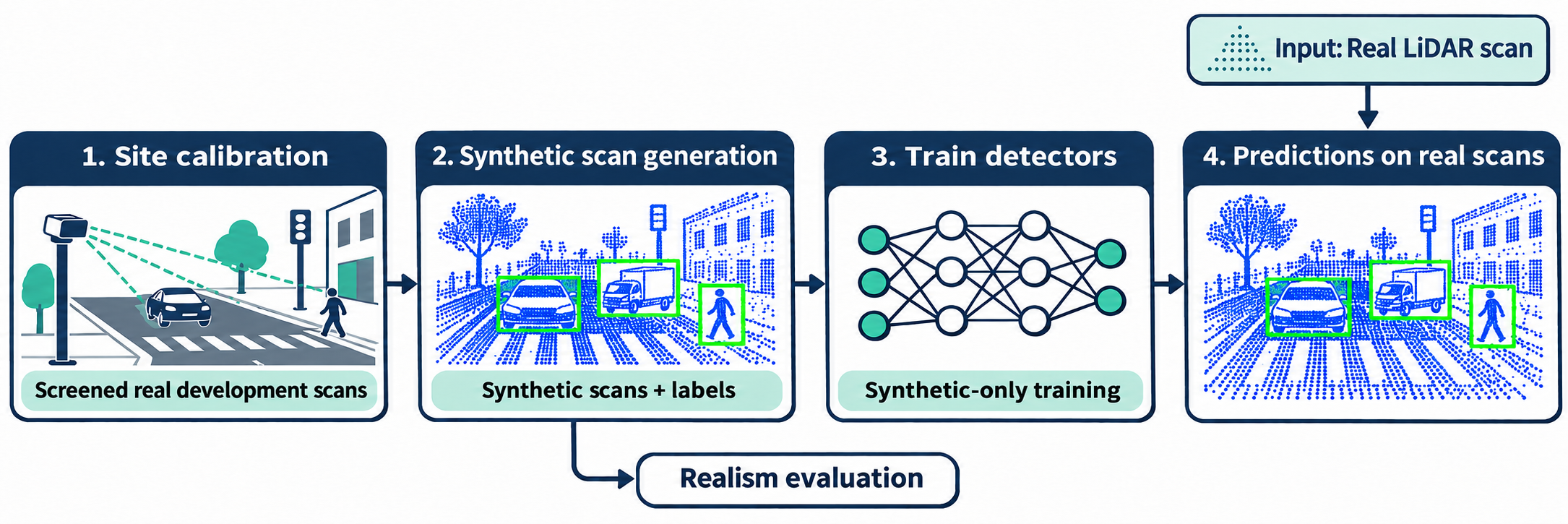}
    \caption{Overview of the proposed pipeline, from site calibration
    and synthetic data generation to detector training and evaluation
    on real roadside LiDAR scans.}
    \label{fig:pipeline}
\end{figure*}

Figure~\ref{fig:pipeline} presents the overall proposed methodology. Initially, unlabeled real development point clouds are used to calibrate the roadside scene and sensor characteristics. These calibrated characteristics are then combined with simulated traffic objects to generate synthetic LiDAR scans and corresponding 3D annotations. Subsequently, multiple 3D object detectors are trained using the synthetic data. During inference, the detectors process real roadside scans, which are aligned to the synthetic ground plane and analyzed at several ground heights. The resulting predictions are aggregated to localize and classify road users. The 50 synthetic scans submitted for evaluation are also generated by the same process and adjusted to match the real sensor's range, enabling assessment of their distributional realism.

The following subsections describe preprocessing, synthetic data generation,  detectors, their training, and inference respectively.

\subsection{Data Preprocessing}

All point clouds are represented within a continuous coordinate system for detector training and evaluation. For V2X Track B, the points are restricted to the range $[-40,-40,-8,40,40,2]$~m. Specifically, $x$ and $y$ coordinates are limited to $[-40,40]$~m, and $z$ coordinates are limited to $[-8,2]$~m. Points outside these bounds are excluded from further processing.

A challenge arises when applying detectors trained on synthetic scans to real scans due to differing input formats. The detectors utilize the four-channel point format $(x,y,z,\text{intensity})$ as specified by the OpenPCDet toolbox. In contrast, real scans contain only coordinate information, and the intensity values in all synthetic training frames are zero. To resolve this discrepancy, a constant zero-intensity channel is appended to each real scan. This keeps the input format consistent with that used during training.

The point cloud is also leveled before the inference to remove the geometric mismatch between tilted roadside sensor and level synthetic ground. The real sensor is tilted by approximately $6.5^\circ$, while the ground in the synthetic training frames is a level plane at $z=-3.0$~m. For each real scan, the ground plane is fitted from that scan’s own points, without labels and without using any other scan. Then the scan is rotated and translated so that the plane is aligned with the synthetic ground level. After inference, the predicted bounding boxes and angles are transformed back to the original coordinate frame for evaluation.

\subsection{Synthetic Data Generation}
\label{sec:synth}

Labeled roadside LiDAR data were produced through a combination of approaches.
These approaches include simulated scenes and a custom site-calibrated pipeline. The simulated scenes are derived from roadside environments generated using the CARLA
simulator and a custom Generator V2.

The sensor's field of view is further divided into small angular regions, and the return of LiDAR values observed in each region is analyzed across unlabeled real development point clouds (where no annotations were used). For each region, the typical distance and direction of returns are estimated, along with the frequency of observed returns. Elevation bins of 0.25$^\circ$ from $-17.5^\circ$ to $17^\circ$ and 1,024 azimuth bins are used. For each cell, the spread of the range and the beam direction are also stored. To prevent parked or stationary objects from being incorporated into the background, points inside the confident boxes predicted by the detector are removed from each scan.

The resulting data are subsequently used to generate the background point cloud that reflects the fixed roadside sensor's view. The same background model also assists in determining which beams do not return. In a real scan, every beam is recorded, and a beam without a return is stored as a point at the origin. With 15,600 points per scan, approximately 31\% of the points within the scoring range are of this type. Both the detector, which receives the points at inference, and the realism metric, which pools all points, are affected by these no-return points.

The modeled background provides the roadside setting but does not include the diversity of road users required for detector training. To address this limitation, labeled traffic scenes are created by adding vehicles, pedestrians, and trucks from previously generated synthetic scenes to the modeled background. Each object is placed on the modeled ground surface, and its position and orientation are sampled. Once an object is placed, its bounding box and class label are transformed along with its
points, thereby providing annotations for detector training. Each synthetic scene contains between 14 and 35 objects.

Although the scene is structured to resemble a real scan, discrepancies remain in the sizes of objects compared to actual road users. For example, synthetic trucks were approximately 8.5 m long and 3.6 m high, whereas real labeled trucks and vans are typically about 6 m long. This issue is addressed by resizing trucks (length 4.6–6.8 m), varying the footprint and height of pedestrians.

During object placement, an additional challenge is to determine how the placed object affects the visibility of background regions from the sensor's perspective. To address this, background points that fall behind an object from the sensor's viewpoint are removed. This process introduces occlusion in the generated scan, ensuring that the detector is trained on scenes where road users can obstruct the surrounding environment.

The 50 synthetic scans submitted for the realism evaluation are generated using the same generator. Because the realism score compares pooled point distributions, these scans are adjusted to match the real sensor's statistics. The proportion of no-return points is set to the real value of 30.8\% within the scoring range, and non-zero points are thinned by range band so that their range histogram aligns with that of the real development scans.

\subsection{Model Architecture}

We train three families of 3D detectors with the OpenPCDet toolbox \cite{openpcdet}, all from scratch on synthetic data:

\begin{itemize}
    \item \textbf{SECOND} \cite{yan2018second} voxelizes the point cloud (voxel size $0.2\times0.2\times0.25$\,m), applies a sparse 3D convolutional backbone, collapses it to a bird's-eye-view feature map and predicts oriented boxes with an anchor-based head.
    \item \textbf{VoxelNeXt} \cite{chen2023voxelnext} is fully sparse: it predicts boxes directly from sparse voxel features without dense maps or anchors, which lets us use a finer $0.1\times0.1\times0.2$\,m grid that helps small objects.
    \item \textbf{PV-RCNN} \cite{shi2020pvrcnn} is a two-stage detector: a voxel-based network proposes boxes, and features pooled from 4,096 keypoints sampled on the raw points refine each proposal.
\end{itemize}

Each family of models is trained separately and no architecture is modified. Besides three-class models (vehicle, pedestrian, truck), we train class-specific specialists, which treat the other classes as background. The detectors are hence independent as one model's output during the training does not affect the other. Our contribution is a framework that combines all the models at the inference level, where each detector only contributes to the classes for which it is validated, and the framework fuses their predictions by clustering box centers, which is further described in Section~\ref{sec:inference}, with its effect on accuracy in Section~\ref{sec:results}.

\begin{table}[H]
\caption{Hyperparameter settings of the three detector families}
\label{tab:hyperparams}
\centering
\footnotesize
\setlength{\tabcolsep}{3.5pt}
\begin{tabular*}{\columnwidth}{@{\extracolsep{\fill}}lccc@{}}
\toprule
 & \textbf{SECOND} & \textbf{VoxelNeXt} & \textbf{PV-RCNN} \\
\midrule
Voxel size (m) & $0.2{\times}0.2{\times}0.25$ & $0.1{\times}0.1{\times}0.2$ & $0.05{\times}0.05{\times}0.25$ \\
Learning rate & 0.003 & 0.001 & 0.002 \\
Batch size & 4 & 2--4 & 4 \\
Epochs & 12 & 12--20$^{a}$ & 4--6 \\
\bottomrule
\end{tabular*}\\[0.6ex]
\begin{minipage}{\columnwidth}
\raggedright\footnotesize
$^{a}$Plus 4--5 epochs of fine-tuning from our own synthetic-trained checkpoints.\\[0.3ex]
All models: one-cycle Adam, mixed precision, trained from scratch on synthetic data.
\end{minipage}
\end{table}

\subsection{Training Procedure}

All detectors are trained from scratch exclusively on synthetic data, employing the standard OpenPCDet objectives: classification loss, smooth-L1 box regression loss on the residuals of the seven box parameters, direction classification loss, as well as the proposal-refinement and point-segmentation losses of PV-RCNN.

The one-cycle Adam optimizer with mixed precision is employed. Learning rates, batch sizes, and epochs are detailed in Table~\ref{tab:hyperparams}. Training scenes are augmented using random flips along the $x$ and $y$ axes, rotations up to $\pm 45^\circ$, scaling between 0.95 and 1.05, and vertical translation noise of 0.15\,m.

Random seeds are not fixed, allowing repeated runs of the model to differ and serve as ensemble members. Several VoxelNeXt models are fine-tuned for a few epochs from synthetic-trained checkpoints on the extended synthetic set. Selection of checkpoints and models for the final ensemble is based on performance on real development sets, as detailed in Section \ref{sec:setup}, and never on the test set.

\subsection{Inference}\label{sec:inference}

Having a single detector pass proved to be unreliable on the real scans for two reasons:

\begin{itemize}
    \item the best ground height differed between models and classes
    \item no single model was best on every class
\end{itemize}

To address these challenges, each leveled scan is processed by every detector at six ground heights ($-3.15$ to $-2.775$\,m). Each (model, height) pair constitutes a single pass, and a model contributes only to the classes for which it is validated.

The boxes from all passes are subsequently fused per class by clustering their bird's-eye-view centers. The resulting fused box is computed as the score-weighted mean of its cluster, with its score defined as the sum of the member scores divided by the number of passes.

\begin{equation}
\mathbf{b} = \frac{\sum_{k} s_k \mathbf{b}_k}{\sum_{k} s_k}, \qquad s = \min\!\Big(\frac{1}{N_{\text{pass}}}\sum_{k} s_k,\; 1\Big),
\end{equation}
where $N_{\text{pass}}$ is the number of passes.

Despite fusion, numerous class-specific errors persisted and were addressed individually. The heading of each fused vehicle and pedestrian box is the score-weighted circular mean of its $K$ members' headings, computed on doubled angles so that a box and its $180^\circ$ flip do not cancel:
\begin{equation}
\psi = \tfrac{1}{2}\operatorname{atan2}\!\Big(\sum_{k=1}^{K} s_k \sin 2\psi_k,\; \sum_{k=1}^{K} s_k \cos 2\psi_k\Big).
\end{equation}

For vehicles only the best of each (model and height) pass counts toward the fused score. Vehicle and pedestrian boxes are calibrated for systematic offsets, and pedestrian scores are lowered by the fraction $f$ of the box occupied by static structures, $s \leftarrow s\,(1-0.8f)$, using a map built from unlabeled development scans.  Truck-like vehicle boxes are also emitted as trucks, duplicate truck boxes are removed, and long trucks receive low-scored size variants. Finally, boxes with few scan points are often false positives. With $n$ the number of non-ground points inside the box enlarged by 0.1\,m,\begin{equation}
s \leftarrow
\begin{cases}
0.25\,s, & n < n_{\min}\\
s, & \text{otherwise,}
\end{cases}
\qquad n_{\min} = \begin{cases} 5 & \text{pedestrian}\\ 30 & \text{truck.}\end{cases}
\end{equation}
These thresholds were chosen on the development sets, and the rule uses only each scan's own points.

To further optimize performance, analysis revealed that approximately one third of the pedestrian boxes were close to the label but too imprecise to achieve an Intersection over Union (IoU) of 0.5. To address this, each pedestrian box generates several size variants and small positional shifts, which are ranked below the original box.

\section{Experimental Setup}
\label{sec:setup}

\subsection{Computing Resources}

Development, scoring, and the majority of detector model training were conducted on a laptop equipped with an AMD RYZEN 9 7940HS CPU, NVIDIA GeForce RTX 4070 GPU, and 16 CPU threads. The PV-RCNN models and class-specific specialists were trained on the Ohio Supercomputer Center (OSC) \cite{ref_osc}, Ascend cluster, utilizing one NVIDIA A100 40 GB GPU, 8 CPU cores, and 32 GB of RAM per job. In total, training on the OSC consumed approximately 47 A100 GPU hours, while the laptop accounted for about 110 GPU hours (approximately 100 for 26 training runs and about 10 for inference), with several runs sharing the laptop GPU concurrently. Synthetic data generation did not require GPU resources.

\subsection{Validation Protocol}

Candidate validation changes were evaluated on two real development sets using the organizers' scoring code, with average precision (AP) reported for each full set and its two alternating halves. A change was implemented only if it improved performance on both sets and all four halves. Additionally, truck AP is reported without including a fixed, truck-sized object that dominates the development labels, as truck AP on the development sets was found to be a poor predictor of leaderboard performance.

\begin{table}[t]
\caption{Official results of our final submission (ID 1054222)}
\label{tab:final}
\centering
\small
\setlength{\tabcolsep}{4pt}
\begin{tabular}{@{}lr@{}}
\toprule
Metric & Value \\
\midrule
\textbf{Combined score} & \textbf{0.8947} \\
\midrule
3D mAP@0.5 (R40) & 0.6001 \\
\quad Vehicle AP & 75.59 \\
\quad Pedestrian AP & 44.95 \\
\quad Truck AP & 59.50 \\
3D mAP@0.7 (R40) & 0.2646 \\
BEV mAP@0.5 (R40) & 0.6430 \\
\midrule
Realism score & 0.9507 \\
\quad Chamfer distance (CD) & 0.6205 \\
\quad MMD & $2.35\times10^{-4}$ \\
\quad Earth mover's distance (EMD) & 0.966 \\
\quad Fréchet point-cloud distance (FPD) & 0.1794 \\
\bottomrule
\end{tabular}
\end{table}

\begin{table}[t]
\caption{Effect of each component on the development sets (150 / 350 frames; AP in \%, realism on the 0--1 scale)}
\label{tab:ablation}
\centering
\footnotesize
\setlength{\tabcolsep}{3pt}
\resizebox{\columnwidth}{!}{%
\begin{tabular}{@{}llcc@{}}
\toprule
Component & Metric & 150 frames & 350 frames \\
\midrule
Per-scan levelling$^{a}$         & mAP       & \multicolumn{2}{c}{15.8$\to$37.2} \\
Multi-height/model fusion$^{a}$  & mAP       & \multicolumn{2}{c}{49.3$\to$56.3} \\
Averaged vehicle heading         & Veh.\ AP  & 75.4$\to$80.0 & 75.4$\to$79.7 \\
Truck de-duplication             & Truck AP$^{c}$ & 66.2$\to$71.0 & 63.7$\to$68.4 \\
Pedestrian hedging               & Ped.\ AP  & 38.5$\to$42.4 & 37.1$\to$41.3 \\
PV-RCNN voters                   & Veh.\ AP  & 82.5$\to$83.5 & 82.4$\to$83.5 \\
Ped.\ specialist (3$\times$)     & Ped.\ AP  & 43.2$\to$44.3 & 42.8$\to$44.0 \\
One vote per pass (veh.)         & Veh.\ AP  & 83.7$\to$84.2 & 83.6$\to$84.0 \\
Averaged ped.\ heading           & Ped.\ AP  & 45.4$\to$46.1 & 45.2$\to$45.4 \\
Low-point ped.\ rule ($<$5 pts)  & Ped.\ AP  & 46.1$\to$46.6 & 45.4$\to$45.9 \\
Low-point truck rule ($<$30 pts)$^{b}$ & Truck AP$^{c}$ & 61.6$\to$62.0 & 61.4$\to$61.6 \\
No-return trimming               & Realism   & \multicolumn{2}{c}{0.9405$\to$0.9488} \\
\bottomrule
\end{tabular}}

\vspace{2pt}
\parbox{\columnwidth}{\scriptsize
$^{a}$ Measured on an earlier single development set; single-seed noise about 3 points.
$^{b}$ Measured on the final ensemble, so not directly comparable with the earlier truck row.
$^{c}$ Excluding a repeated parked truck in the development labels (Section~\ref{sec:setup}).}
\end{table}

\section{Experimental Results}
\label{sec:results}

\subsection{Main Results}
Table~\ref{tab:final} presents the official results of the final submission. The method achieved a combined score of 0.8947, with a 3D mAP@0.5 of 0.6001 and a realism score of 0.9507, and was ranked joint first on the official final leaderboard~\cite{ref_challenge}.
According to~\eqref{eq:combined_score}, the normalized detection term (0.8573) contributes 0.514 to the overall score, while the realism component contributes 0.380.

In the final version, vehicles represent the strongest class (75.59 AP). This accuracy primarily results from fusion, which combines several detectors at multiple ground heights and averages headings across the fused boxes, producing stable vehicle boxes even when individual passes disagree.

Pedestrians remained the most challenging class (44.95 AP) because many are represented by only a few points, resulting in boxes that are often close to the label but too loose to be counted at an IoU of 0.5. Size-graded hedging and the low-point rule (Section~\ref{sec:inference}) address this issue: hedging provides each pedestrian box with several size and position variants, while the low-point rule lowers boxes with too few points to be considered real.

The truck class ranks between the two other classes with 59.50 AP. Its performance primarily depends on the truck size rule, which assists in classifying large vehicle boxes as trucks. However, this class was the most difficult to validate reliably because a repeatedly parked truck inflated truck AP on the development sets (Section~\ref{sec:setup}).

Beyond detection, the submitted scans achieved a realism score of 0.9507 by matching the real sensor's fraction of no-return points, which constitutes approximately 31\% of a real scan. A stricter metric further demonstrated the placement accuracy of detected boxes: at an IoU of 0.7, the 3D mAP of 0.2646 retains about 44\% of the IoU 0.5 performance.

These results indicate that a substantial share of correct detections are also tightly localized, while precise placement of small objects, particularly pedestrians, remains the primary limitation.

\subsection{Ablation Study}

The contribution of each component was quantified by measuring the total effect on the two real development sets, sequentially adding each component to the preceding pipeline in Table~\ref{tab:ablation}.

The most substantial gains resulted from per-scan ground leveling and multi-height, multi-model fusion followed by averaging the headings of vehicle boxes and hedging the pedestrian boxes. Further improvements were achieved by removing duplicate truck boxes, trimming the no-return points of the submitted scans, and incorporating PV-RCNN voters. The gains from PV-RCNN voters were smaller but consistent across both sets.

The development gains associated with the first three components were also reflected on the leaderboard. Truck changes transferred unreliably because development truck AP was inflated by a repeated parked object. The one exception was the low-point truck rule, which raised test truck AP from 59.06 to 59.50 (Section~\ref{sec:results}).


\section{Conclusion}
\label{sec:conclusion}
We propose an inference-level ensemble model for roadside LiDAR 3D object detection, constrained to training exclusively on synthetic data. Our method combines bounding boxes from SECOND, VoxelNeXt, and PV-RCNN detectors, each trained solely on synthetic scans generated by a CARLA-based, site-calibrated pipeline that replicates the fixed sensor’s range image and no-return beams. During inference, each real scan is aligned to the synthetic ground plane and processed by each detector at multiple ground heights. The resulting bounding boxes are fused using bird’s-eye view (BEV) clustering and further refined through class-specific post-processing. Our ensemble achieved a final score of 0.8947, ranked jointly 1st on the final leaderboard, and obtained a realism score of 0.9507. Ablation studies demonstrate that per-scan ground leveling, averaging vehicle headings, and size-graded pedestrian hedging were the most significant contributors.

Several limitations are identified in this approach. The development set average precision (AP) did not predict the leaderboard result because a repeated parked object inflated the metric, preventing validation of truck-related changes; consequently, truck detection could not be tuned reliably. Additionally, pedestrian detection remains the weakest class, as many pedestrians are represented by only a few points, resulting in errors primarily in box placement rather than detection. While size-graded hedging reduces these errors, it does not eliminate them.
The proposed winning solution establishes a strong baseline for bridging the gap between synthetic and real LiDAR data for 3D object detection in Urban V2X scenarios. 

Our future work will focus on the following items:

\begin{itemize}
    \item Develop a validation set that more reliably predicts truck and realism scores
    \item Enhance pedestrian localization using a localization-aware training loss or a learned box-refinement stage
    \item Replace hand-tuned fusion and post-processing rules with a learned fusion model
    \item Evaluate whether the site calibration transfers to other roadside sensors with reduced real data
\end{itemize}

\section*{Code Availability}
The code and configuration files used in this work are publicly available at
\url{https://github.com/Arnavvvkumar/IEEE-BigData-2026-Cup---V2X-Track}.
The repository contains the synthetic data generation pipeline, detector configurations, training scripts, the inference and fusion pipeline, the trained weights of all models in the final ensemble, and the files of the final submission. The training datasets are not included but can be regenerated with the released code.

\section*{Acknowledgment}

This paper was invited as the winning solution of the UCF UrbanTwin Sim2Real LiDAR Challenge at the IEEE BigData 2026 Cup. This work was supported in part by an allocation of computing time from the Ohio Supercomputer Center (OSC) \cite{ref_osc}. Generative AI tools were also used in this work. ChatGPT (OpenAI) was used to generate the illustrations in Fig.~\ref{fig:pipeline} based on the authors' description. Claude (Anthropic) was used as a coding assistant to optimize model training. Grammarly was used for grammar and wording corrections. The authors reviewed and verified all AI-assisted content and take full responsibility for this paper.

\bibliographystyle{IEEEtran}
\bibliography{references}

\end{document}